\documentclass{article}

\usepackage{microtype}
\usepackage{graphicx}
\usepackage{subcaption}
\usepackage{booktabs} 

\usepackage{hyperref}

\usepackage[preprint]{icml2026}

\usepackage{amsmath}
\usepackage{amssymb}
\usepackage{mathtools}
\usepackage{amsthm}

\usepackage[capitalize,noabbrev]{cleveref}

\theoremstyle{plain}

\theoremstyle{definition}

\theoremstyle{remark}

\usepackage{inconsolata}
\usepackage{enumitem}

\usepackage{xspace}
\DeclareRobustCommand{\Sombrero}{\textsc{Sombrero}\xspace}

\usepackage[disable,textsize=tiny]{todonotes}

\icmltitlerunning{SOMBRERO: Measuring and Steering Boundary Placement in End-to-End Hierarchical Sequence Models}

\begin{document}

\twocolumn[
  \icmltitle{SOMBRERO: Measuring and Steering Boundary Placement in End-to-End Hierarchical Sequence Models}



  \icmlsetsymbol{equal}{*}

  \begin{icmlauthorlist}
    \icmlauthor{Pit Neitemeier}{aa}
    \icmlauthor{Alessio Serra}{aa}
    \icmlauthor{Jiaze Li}{aa}
    \icmlauthor{Sascha Wirges}{aa}
    \icmlauthor{Lukas Balles}{aa}
    \icmlauthor{Jan Hendrik Metzen}{aa}

  \end{icmlauthorlist}

  \icmlaffiliation{aa}{Aleph Alpha Research}

  \icmlcorrespondingauthor{Pit Neitemeier}{pit.neitemeier@aleph-alpha-research.com}
  \icmlcorrespondingauthor{Jan Hendrik Metzen}{janmetzen@mailbox.org}

  \icmlkeywords{Machine Learning, ICML}

  \vskip 0.3in
]



\printAffiliationsAndNotice{}  

\begin{abstract}
Hierarchical sequence models replace fixed tokenization with learned segmentations that compress long byte sequences for efficient autoregressive modeling. While recent end-to-end methods can learn meaningful boundaries from the language-modeling objective alone, it remains difficult to \emph{quantitatively} assess and \emph{systematically} steer where compute is spent.
We introduce a router-agnostic metric of \emph{boundary quality}, \emph{boundary enrichment} $B$, which measures how strongly chunk starts concentrate on positions with high next-byte surprisal.
Guided by this metric, we propose \Sombrero, which steers boundary placement toward predictive difficulty via a \emph{confidence-alignment boundary loss} and stabilizes boundary learning by applying confidence-weighted smoothing at the input level rather than on realized chunks.
On 1B scale, across UTF-8 corpora covering English and German text as well as code and mathematical content, \Sombrero improves the accuracy--efficiency trade-off and yields boundaries that more consistently align compute with hard-to-predict positions.
\end{abstract}

\section{Introduction}
Tokenization is a pragmatic compression layer for language modeling \citep{gage1994bpe,sennrich2016bpe,kudo2018sentencepiece,radford2019gpt2}, but it is also a hard-coded interface between data and model:
it fixes a discrete vocabulary \citep{clark2022canine},
bakes in language- and domain-specific heuristics \citep{kudo2018sentencepiece,radford2019gpt2},
and can yield brittle behavior around spelling, morphology, and character-level perturbations \citep{pruthi2019misspellings,geh2025adversarialtokenization}.
\emph{Hierarchical sequence models} sidestep these issues by allowing models to operate directly on \emph{byte sequences} \citep{xue2022byt5},
enabling a single model to cover natural language, code, and mathematical text without changing preprocessing.
This unification removes tokenization as a modeling choice, but shifts the burden of efficiency and abstraction entirely onto the model architecture.
The main obstacle is efficiency: byte-level sequences are substantially longer than subword token sequences, so na\"ively applying isotropic Transformers at byte level is compute-inefficient \citep{xue2022byt5,clark2022canine}.
\emph{Hierarchical Autoregressive Transformers} (HATs\footnote{We adopt the term Hierarchical Autoregressive Transformer from \citet{neitemeier2025hat}; here, we use it to denote the broader family of methods that take byte-level sequences as input and output, while internally pooling to a coarser, token-like granularity.}) address this by acting as \emph{adaptive compute allocators}: a lightweight local model processes every byte, while a heavier backbone is invoked on a compressed representation.

The effectiveness of a HAT hinges on \emph{where} it places chunk boundaries, determining where expensive backbone compute is spent.
Yet, for end-to-end learned chunking it remains difficult to (i) \emph{measure} whether boundaries target predictive difficulty and (ii) \emph{control} boundary placement to improve the accuracy--efficiency trade-off.

Existing approaches largely fall into two families. \emph{Fixed-chunk hierarchical architectures} use predefined segmentation rules or chunk sizes to reduce sequence length,
for example by processing fixed byte chunks with local modules and a global backbone \citep{yu2023megabyte},
by injecting larger blocks only at special boundary bytes such as spaces \citep{slagle2024spacebyte},
or by combining a byte-level encoder/decoder with a word-level backbone via a fixed notion of higher-level units \citep{neitemeier2025hat}.
An alternative design is to derive chunk boundaries from auxiliary models, e.g., using the entropy of a small byte-level model as a segmentation signal \citep{pagnoni2024blt}.
While these approaches can be simple and efficient, they hard-code where resolution changes occur and decouple boundary placement from the end-to-end training objective.

In contrast, \emph{learned-chunk hierarchical architectures} infer boundaries end-to-end from data.
Recent work (H-Net) shows that learned segmentation can close the gap to tokenized models while improving robustness to character-level variation \citep{hwang2025hnet};
subsequent efforts demonstrate that existing subword models can be ``byteified'' using learned boundary mechanisms \citep{minixhofer2025bolmo}.
H-Net, in particular, combines a cosine-similarity-based router, chunk-level confidence-weighted smoothing, and an auxiliary ratio loss targeting a desired compression ratio. \citet{hwang2025hnet} evaluate boundary structure primarily through qualitative analyses. 

\textbf{First}, we introduce a router-agnostic metric for \emph{boundary quality} based on a simple operational principle: boundaries should be placed at positions where predicting the next byte is hard, warranting additional backbone compute.
Concretely, we measure a \emph{boundary enrichment} score $B$, defined as the ratio between the average next-byte surprisal at chunk starts and the average surprisal over all positions.
We verify that enrichment is significant under a rate-matched circular-shift null and show that $B$ aligns with end-to-end model performance across domains and evaluation settings.

\textbf{Second}, guided by this lens, we introduce an auxiliary \emph{boundary-shaping loss} that explicitly steers boundary placement toward predictive difficulty.
Our \emph{confidence-alignment boundary loss} aligns boundary scores with per-byte target confidence, encouraging backbone steps to focus on high-surprisal bytes.
Unlike entropy-based chunking of \citet{pagnoni2024blt}, which derives boundaries from a separate predictive signal, our loss is optimized jointly with the language modeling objective in an on-policy manner.

\textbf{Third}, compared to H-Net, we move confidence-weighted temporal smoothing from realized chunks to the underlying byte-level sequence, ensuring dense gradient flow to every boundary score rather than only those that cross a discretization threshold.
Moreover, our variant operates at equidistant inputs (rather than variable-length realized chunks), reducing early-training instability and enabling a simpler boundary parameterization (a linear projection with sigmoid probabilities rather than cosine similarity between adjacent byte representations).

We denote the resulting architecture as \Sombrero, a special HAT.
Across experiments at 1B scale on UTF-8 byte sequences spanning English and German natural language as well as code and mathematical text,
\Sombrero consistently improves accuracy--efficiency trade-offs while producing boundaries that better align with predictive difficulty.
We conclude that these results suggest that learned chunking is an effective mechanism for \emph{adaptive compute allocation} in hierarchical sequence models.

\begin{figure*}[tb]
\includegraphics[width=.9\textwidth]{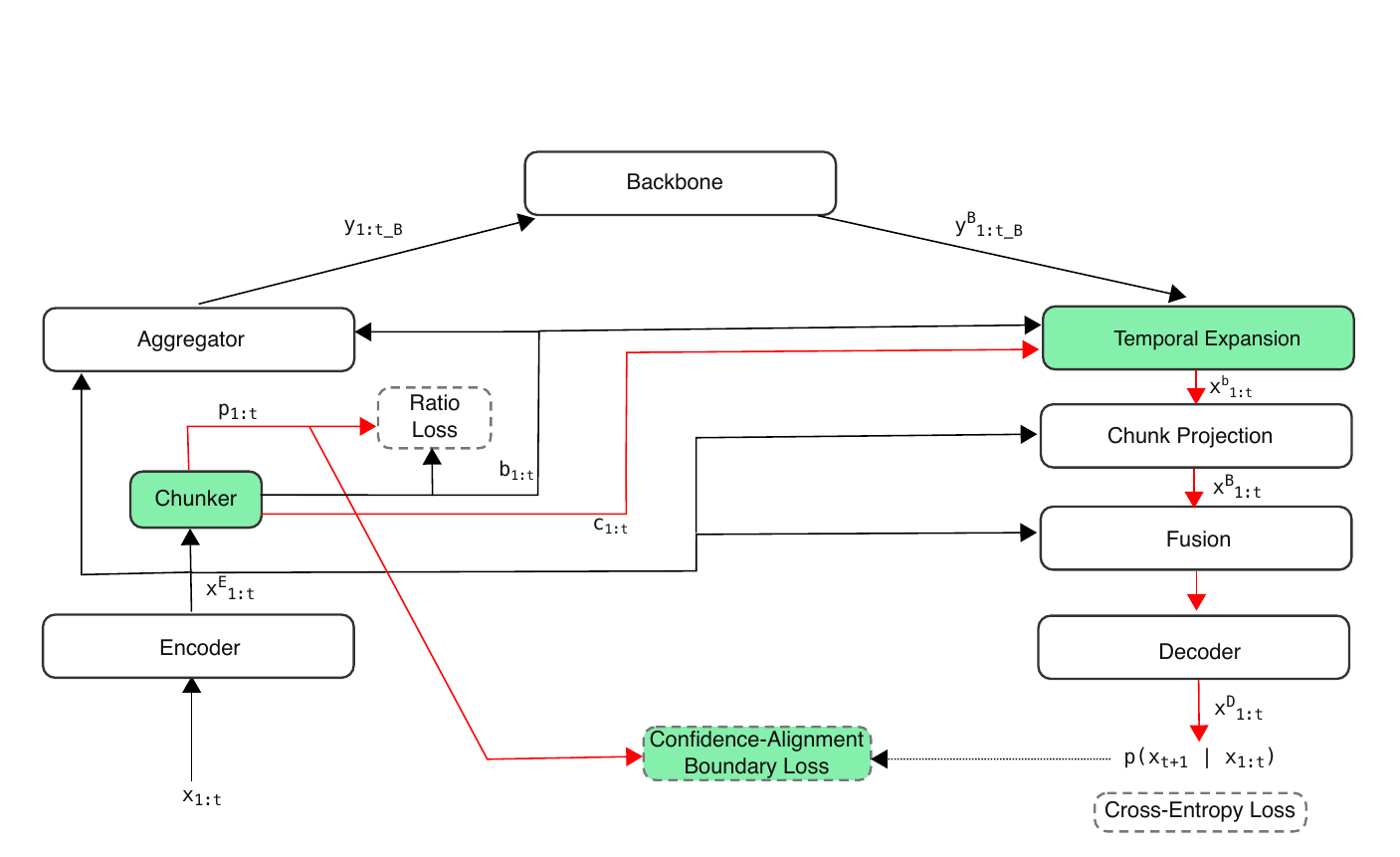}
\caption{Illustration of the HAT architecture and information flow. Arrows in red indicate connections along which gradient affecting the boundaries $\mathbf{b}$ flows backward. Green boxes indicate the changes contributed in \Sombrero: a simplified sigmoid chunker, byte-level temporal expansion, and an additional confidence-alignment boundary loss.}
\label{figure:hat_architecture}
\end{figure*}

\section{Related Work}
\label{sec:related_work}

Most modern language models (LMs) rely on a learned subword tokenizer---e.g., BPE \citep{sennrich2016bpe}, WordPiece \citep{schuster2012wordpiece}, or SentencePiece \citep{kudo2018sentencepiece}---which constructs a discrete vocabulary prior to training.
Recent work also revisits the tokenizer design itself; for example, SuperBPE extends BPE to learn ``superword'' units that can span whitespace, improving efficiency without removing tokenization \citep{liu2025superbpe}.
Hierarchical BPE explores dynamic grouping on top of BPE structure, bridging tokenization and internal segmentation by inducing hierarchical groupings during training \citep{dolga2025hierarchicalbpe}.
Tokenizer-free approaches aim to remove this extraneous step and instead operate on raw characters or bytes as inputs and outputs.

A direct way to eliminate tokenization is to apply a standard Transformer on byte or character sequences, relying on scale to learn composition \citep[e.g., ByT5,][]{xue2022byt5}. MambaByte explores replacing attention with selective state-space models to change the inductive bias for long sequences while remaining token-free at the byte level \citep{wang2024mambabyte}. EvaByte studies efficient designs for byte-level LMs at scale, emphasizing architectural and attention-efficiency choices in the tokenizer-free regime \citep{evabyte2025}.
However, several works have observed that it is highly beneficial to focus a model's processing power on a coarser, token-like granularity by some form of internal downsampling \citep{clark2022canine,yu2023megabyte,slagle2024spacebyte,pagnoni2024blt,tay2021charformer,kallini2024mrt5,hwang2025hnet}.
This downsampling can be ``static'' or use a semantically meaningful segmentation, which can be provided externally or learned by the model itself.

\textbf{``Static'' downsampling.}
CANINE operates on characters but internally downsamples via strided convolutions \citep{clark2022canine}.
MEGABYTE downsamples by a simple concatenation over multiple subsequent byte positions \citep{yu2023megabyte}.
Multiscale Byte Language Models (MBLM) explore a closely related multi-rate hierarchy for causal million-length modeling, pairing byte-level processing with coarser representations \citep{egli2025mblm}.

\textbf{Externally-provided segmentations.}
These methods rely on an externally-provided \emph{segmentation} of text into semantically meaningful units (e.g. words) while processing raw bytes/characters without relying on a fixed vocabulary.
CharacterBERT replaces subword embeddings with character-based word representations \citep{elb2020characterbert}.
SpaceByte \citep{slagle2024spacebyte} and HAT \citep{neitemeier2025hat} use a word-like segmentation and process text in a hierarchical fashion.
BLT uses a segmentation derived from the entropy of a separately-trained, byte-level model \citep{pagnoni2024blt}. 
AU-Net constructs a deeper hierarchy by pooling bytes into increasingly coarse units derived from word/group structure \citep{videau2025aunet}.

\textbf{Learned segmentations.}
Charformer introduces a differentiable subword induction mechanism (GBST) that softly selects variable-length blocks and feeds the resulting latent sequence to a Transformer \citep{tay2021charformer}.
MrT5 follows the same motivation as ByT5 but learns to shorten intermediate representations via token deletion/merging inside the encoder \citep{kallini2024mrt5}.
FLEXITOKENS learns byte-level boundaries to form variable-length segments and emphasizes adaptability of the learned segmentation during distribution shift and finetuning \citep{owodunni2025flexitokens}.

\textbf{End-to-end learned dynamic chunking (H-Net).}
H-Net pushes the above ideas toward minimal external bias by learning chunk boundaries jointly with the LM objective and composing them into nested hierarchies, enabling context-dependent compression rather than extraneous segmentation \citep{hwang2025hnet}.
H-Net++ explores refinements of this dynamic-chunking paradigm in morphologically rich settings, highlighting that learned segmentation can incorporate language-specific orthographic cues while remaining tokenizer-free \citep{zakershahrak2025hnetpp}.
Among recent tokenizer-free architectures, end-to-end learned chunking arguably offers the broadest hypothesis class for “what a token could be,” suggesting the strongest potential to benefit from scale when paired with efficient sequence backbones and training compute.

\textbf{Retrofitting and ``byteifying'' token LMs.}
Instead of training tokenizer-free LMs from scratch, Bolmo converts a pretrained subword LM into a byte-level architecture by distilling from the source model, effectively inheriting its higher-level abstractions while removing the fixed subword vocabulary \citep{minixhofer2025bolmo}. This retrofitting perspective is also reflected in domain-specific byte modeling, where non-text byte streams motivate end-to-end tokenization removal \citep{li2025bytegen}.

\section{SOMBRERO}
This section introduces the HAT framework (Section~\ref{subsection:background}) and revisits the chunker parameterization (Section~\ref{subsection:chunker}). Building on these components, we present \textit{byte-level smoothing} for temporal expansion (Section~\ref{subsection:byte_smoothing}) and the \textit{confidence-alignment boundary loss} (Section~\ref{subsection:confidence_alignment_loss}). We refer to the resulting architecture as \Sombrero. We conclude by introducing the \emph{boundary enrichment metric} for quantifying the quality of sequence chunking (Section~\ref{subsection:boundary_enrichment_metric}).

\subsection{Background} \label{subsection:background}
\paragraph{Hierarchical Autoregressive Transformer (HAT)}
We summarize the general framework of \emph{Hierarchical Autoregressive Transformers} (HAT), illustrated in Figure \ref{figure:hat_architecture}. A HAT models the joint distribution over a variable-length sequence $x_{1:T}$ via the standard autoregressive factorization $p(x_{1:T}) = \prod_{t=1}^{T} p(x_t \mid x_{1:t-1}) $. It therefore suffices to describe how a HAT computes $p(x_{t+1} \mid x_{1:t})$.

A HAT consists of three autoregressive subnetworks (deeper nested hierarchical variants are possible; we focus on two levels for clarity and simplicity): an \emph{encoder} $\mathcal{E}$, a \emph{backbone} $\mathcal{B}$, and a \emph{decoder} $\mathcal{D}$. Two connector modules mediate information flow between these subnetworks: the \emph{backbone connector} $\mathcal{C}_\mathcal{B}$ and the \emph{decoder connector} $\mathcal{C}_\mathcal{D}$. Each subnetwork may itself be instantiated by a decoder-style Transformer, Mamba-style recurrent layers, or any other architecture suitable for autoregressive modeling. Different HAT variants primarily differ in the design of the connector layers and in the auxiliary losses used to supervise the backbone connector. We denote intermediate representations at full sequence resolution (length $t$) by $x$ and at reduced sequence resolution  (length $t_\mathcal{B}$) as processed by the backbone by $y$. We denote outputs of a subnetwork by the respective subnetwork's symbol as superscript, e.g., $x^\mathcal{E}$ for the encoder's outputs.

\paragraph{Backbone Connector: Sequence Compression}
Let $x^\mathcal{E}_{1:t}=\mathcal{E}(x_{1:t})$ denote the encoder representations, with $x^\mathcal{E}_i \in \mathbb{R}^{d_o}$. The backbone connector $\mathcal{C}_\mathcal{B}$ performs \emph{sequence compression}. It outputs:
\begin{itemize}[noitemsep,topsep=0pt]
    \item compressed sequence $y_{1:t_\mathcal{B}}$ with $y_j \in \mathbb{R}^{d_i}$ and $t_\mathcal{B} \leq t$
    \item boundary scores $p_{1:t}$ with $p_i \in [0, 1]$
    \item boundary indicators $b_{1:t} \in \{0, 1\}$ with $\sum_i b_i = t_\mathcal{B}$
    \item confidence vectors $c_{1:t}$ with $c_i \in [0,1]$
\end{itemize}
The indicator $\mathbf{b}$ specifies which encoder positions are retained: if $b_i = 1$ is the $j$-th active position in $\mathbf{b}$ , then $y_j$ is derived solely from $x^\mathcal{E}_{1:i}$ and is independent of $x^\mathcal{E}_{i+1:t}$. Conceptually, $\mathcal{C}_\mathcal{B}$ decomposes into (a) a \emph{chunker} that predicts $\mathbf{p}$ from $x^\mathcal{E}_{1:t}$ and derives $\mathbf{b}$ and $\mathbf{c}$ from $\mathbf{p}$, and (b) an \emph{aggregator} that constructs $y_{1:t_\mathcal{B}}$ from $x^\mathcal{E}_{1:t}$ and $b$. The aggregator typically increases feature dimensionality ($d_i > d_o$) to provide the backbone with a richer, more abstract representation.

\paragraph{Decoder Connector: Sequence Expansion and Fusion}
Let $y^\mathcal{B}_{1:t_\mathcal{B}} = \mathcal{B}(y_{1:t_\mathcal{B}})$ denote the backbone outputs, with $y^\mathcal{B}_j \in \mathbb{R}^{d_i}$. The decoder connector $\mathcal{C}_\mathcal{D}$ performs the inverse operation of $\mathcal{C}_\mathcal{B}$: \emph{sequence expansion and fusion}. It produces a sequence $x^\mathcal{D}_{1:t}$ with $x^\mathcal{D}_i \in \mathbb{R}^{d_o}$ and fuses it with the encoder features $x^\mathcal{E}_{1:t}$ (a simple instantiation of fusion is residual addition).
Temporal dependencies must respect the causal compression structure: if $b_i = 1$ corresponds to the $j$-th active element, then $x^\mathcal{D}_i$ may only depend on $y^\mathcal{B}_{1:j}$ and must be independent of $y^\mathcal{B}_{j+1:t_\mathcal{B}}$. Operationally, $\mathcal{C}_\mathcal{D}$ consists of
(a) \emph{temporal expansion} that maps $(\mathbf{y^\mathcal{B}}, \mathbf{b}, \mathbf{c})$ to an expanded sequence $x^b_{1:t}$ with $x^b_i \in \mathbb{R}^{d_i}$, 
(b) \emph{chunk projection} from $(x^\mathcal{E}, x^b)$ to decoder-space features $x^\mathcal{B}$, and
(c) \emph{fusion} that combines $x^\mathcal{B}$ with $x^\mathcal{E}$ to form the input to the decoder $\mathcal{D}$.

\begin{figure*}[tb]
\includegraphics[width=.95\linewidth]{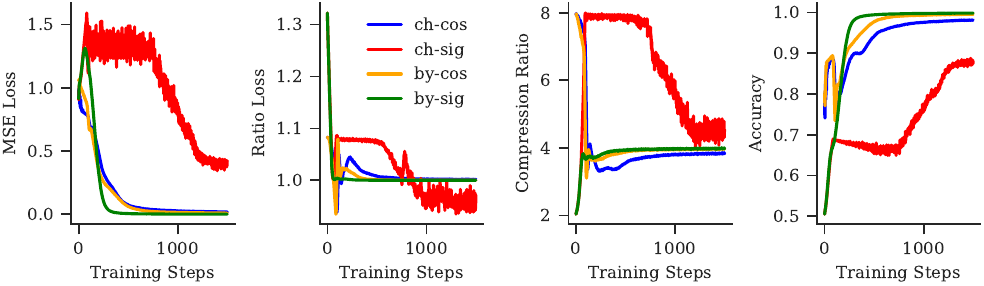}
\caption{We compare \textbf{ch}unk- and \textbf{by}te-level smoothing in combination with \textbf{cos}ine and \textbf{sig}moid boundary scores on a synthetic problem (see Section \ref{appendix:synthetic_problem}). Sigmoid boundary scores underperform with chunk-level smoothing but excel with byte-level smoothing.}
\label{figure:synthetic_problem}
\end{figure*}

\subsection{Chunker} \label{subsection:chunker}
H-Net predicts boundary scores $\mathbf{p}$ by linearly projecting $x^\mathcal{E}_i$ to queries $q_i = W_q x^\mathcal{E}_i$ and keys $k_i = W_k x^\mathcal{E}_i$, and then defining boundary scores via the \emph{cosine} similarity of adjacent queries and keys: $p_i = \frac{1}{2}\Bigl(1 - \frac{q_i^\top k_{i-1}}{|q_i|\cdot |k_{i-1}|}\Bigr)$ with the convention $p_1 = 1.0$.
Based on these boundary scores, H-Net derives boundaries by thresholding $b_i = \mathbf{1}_{{p_i > 0.5}}$ and confidences $c_i= \max(p_i, 1-p_i)$.
H-Net then uses the ratio loss to encourage that the empirical compression rate $C_{emp} = t / t_\mathcal{B}$ matches a target compression ratio $C_{tar}$. 

\citet{hwang2025hnet} motivate the cosine-similarity parametrization by arguing that it ``adds an inductive bias by measuring the similarity between adjacent representations: when context changes, consecutive vectors should exhibit lower similarity'', and show that removing it degrades performance.
We offer an alternative explanation: for high-dimensional, randomly initialized representations, cosine similarities concentrate around $0$, so $p_i$ concentrates near $0.5$.
Driving scores toward $0$ or $1$ then requires precise query/key alignment and thus becomes feasible mainly after the encoder learns meaningful features.
A simpler linear \emph{sigmoid} predictor ($p_i = \mathrm{sigmoid}(W\cdot x^\mathcal{E}_i)$) lacks this \emph{``delayed learning''} effect; we hypothesize that the delay can be helpful when early training dynamics otherwise drift toward over- or under-compression.

To test this, we study a controlled \emph{synthetic change-point task}: sequences are generated by a piecewise-constant latent process with random change points and observed via a noisy linear projection (details in Appendix~\ref{appendix:synthetic_problem}).
Figure~\ref{figure:synthetic_problem} compares cosine vs. sigmoid parameterizations under H-Net's default chunk-level temporal expansion.
The sigmoid variant initially overcompresses (compression is capped at 8.0) and exhibits high MSE, recovering only slowly and unstably, whereas the cosine variant avoids this degeneracy (with the expected slower learning in the first 100 steps).
We next revisit temporal expansion and how it can incentivize overcompression.

\subsection{Byte-Level Smoothing} \label{subsection:byte_smoothing}
Temporal expansion maps $(\mathbf{y^\mathcal{B}}, \mathbf{b}, \mathbf{c})$ to an expanded sequence $x^b_{1:t}$ with $x^b_i \in \mathbb{R}^{d_i}$. H-Net applies confidence-based smoothing via an exponential moving average on the \emph{chunk}-level representations: $ \bar{y}_{i} = c_{T(i)}\cdot y^\mathcal{B}_i + (1 - c_{T(i)})\cdot \bar{y}_{i-1} $, where $T(i)$ is the index of the $i$-th active entry in $b$. Subsequently, $\mathbf{\bar{y}}$ is expanded by repetition (each $\bar{y}_i$ is repeated $T(i+1) - T(i)$ times) to form $\mathbf{x^b}$. Note that this smoothing (i) ignores all confidences $c_j$ for non-boundary positions $b_j \neq 1$ and (ii) does not account for the varying expansion ranges $T(i+1) - T(i)$. We argue that this can bias the initial learning signal: there is a phase during early training where backbone features are low signal and high variance and long-horizon temporal smoothing is beneficial, which drives confidences $\mathbf{c}$ and the corresponding boundary scores $\mathbf{p}$ down (lower confidences correspond to longer horizon). This in turn can result in fewer chunk boundaries and an \emph{``overcompression bias''} in early training as evidenced  with the sigmoid chunker in Figure \ref{figure:synthetic_problem}.

To avoid this bias in early learning, we propose \emph{byte}-level smoothing: we first temporally expand the sequence $y^\mathcal{B}_{1:t_\mathcal{B}}$ to $x^k_{1:t}$ by repetition (each $y^\mathcal{B}_i$ is again repeated $T(i+1) - T(i)$ times). Subsequently, we apply confidence-based smoothing on the extended sequence via $x^b_i = c_i\cdot x^k_i + (1 - c_i) \cdot x^b_{i-1}$. This has the advantage that it takes into account all entries of $c_i$, thus also reducing confidences (and increasing boundary scores) of non-boundaries and balancing training signal for the boundary scores $\mathbf{p}$. Moreover, byte-level smoothing operates over temporally equidistant entries, which is more sound conceptually. This comes at the expense of slightly increased (but overall insignificant) compute because the sequence length of temporal smoothing increases from $t_\mathcal{B}$ to $t$. As shown in Figure \ref{figure:synthetic_problem} this byte-level smoothing accelerates and stabilizes learning on the synthetic problem from Section \ref{subsection:chunker}: MSE increases only briefly, compression does not overshoot, and the sigmoid quickly recovers and surpasses the cosine parameterization; the cosine also benefits, but less.

We note that H-Net reintroduces per-byte confidences during fusion of encoder and backbone outputs. More specifically, before adding upsampled backbone features $x^\mathcal{B}$ with $x^\mathcal{E}$, the $x^\mathcal{B}$ are multiplied with the corresponding boundary confidences, while applying a straight-through estimator \cite{2013arXiv1305.2982B} to the confidence. We argue (and show empirically) that this way of reintroducing confidences is not required with byte-level smoothing and simply adding encoder features and upsampled backbone features is sufficient.

\subsection{Confidence-Alignment Boundary Loss} \label{subsection:confidence_alignment_loss}
While H-Net demonstrated that end-to-end learning of sequence chunking is feasible, we hypothesize that auxiliary losses that directly supervise the boundary scores can accelerate and improve learning of effective chunking. Specifically, we build on the intuition that learned chunking should align with natural sequence segments, such as (sub-)words in natural text, symbols and operators in math and code, or numbers. Naturally, in many cases within-segment predictions are easier than next-segment prediction, that is: the first byte of a new segment is typically more difficult to predict than other bytes (for instance, completing the last character of a word is often straightforward). We introduce a metric for quantifying this property in Section \ref{subsection:boundary_enrichment_metric}.

Following this intuition, we propose to add a \emph{confidence-alignment boundary (CAB)} loss. This loss directly supervises boundary scores $p_t$ based on the hypothesis that $p_t$ should be high at positions where the model has high surprisal on the next-byte target (i.e., assigns low probability). Specifically, let $P_{t+1} = p_\theta(x_{t+1}\vert x_{1:t})$ be the probability assigned by the model $p_\theta$ to the next byte $x_{t+1}$ given context $x_{1:t}$, with $-\log P_{t+1}$ being $x_{t+1}$'s surprisal. We then define the CAB loss as $(1-\mathbf{sg}[P_{t+1}] - p_t)^2$, where $\mathbf{sg}$ denotes the stop-gradient operation. We match $p_t$ to $1 - P_{t+1}$ (rather than the monotonically related surprisal) because $P_{t+1}\in[0,1]$ lives in the same interval as the boundary scores. In practice, we clamp both $P_{t+1}$ and $p_t$ to the range $[10^{-6}, 1 - 10^{-6}]$.

We note that the CAB loss is not (but could be) based on an auxiliary model: our instantiation of the CAB loss supervises boundary scores by the same model's predictions, where model predictions in turn depend on the current chunking that results from the boundary scores. While circular, this has the benefit of not requiring any additional forward passes of auxiliary models and avoiding mismatch of targets between auxiliary and actual model due to being ``on-policy''. We also point out that while the CAB loss supervises the boundary scores directly, it does not determine them on its own but in conjunction with end-to-end gradients coming from the cross-entropy loss on the actual next-byte prediction task via the temporal expansion (see Figure \ref{figure:hat_architecture} for an illustration of the gradient-flow paths affecting the chunker).

\subsection{Boundary Enrichment Metric} \label{subsection:boundary_enrichment_metric}
To quantify whether a learned segmentation places chunk starts at locations that can benefit from the extra backbone compute, we measure how strongly boundaries concentrate on high-surprisal bytes. Let $s_{t+1}=-\log p(x_{t+1}\mid x_{1:t})$ denote the next-byte surprisal, and let $b_t\in\{0,1\}$ indicate a chunk start at position $t$. We score each potential boundary by a local hardness signal $h_t=s_{t+1}$. The \emph{boundary enrichment} is the normalized ratio
$
B \;=\; (\frac{1}{\sum_t b_t}\sum_t b_t\,h_t) / (\frac{1}{T}\sum_t h_t),
$
where $B>1$ indicates that boundaries preferentially occur at high-hardness positions. We also ensure that results are significant under a rate-matched circular-shift null: we compute a null distribution by circularly shifting the boundary indicator sequence (preserving the exact boundary pattern up to rotation) and check a $Z$-score $Z_B=(B-\mu_{\text{null}})/\sigma_{\text{null}}$, where $\mu_{\text{null}},\sigma_{\text{null}}$ are the mean and standard deviation under the null.

\section{Experiments}
We conduct a sequence of experiments that form a \emph{design ladder}: we start from a strong baseline (H-Net) and gradually revisit core parts of architecture, loss, and metrics.

\begin{table*}[tb]
\caption{SOMBRERO Design Ladder Results with bits-per-byte (BPB) averaged globally and per domain (English, German, Code, and Math). Boundary statistics include the boundary enrichment metric $B$, the empirical compression rate $C_{emp}$, normalized gap entropy $H_g$, CUSUM range $R_{\mathrm{CUSUM}}$, and runs test z-score $Z_{\mathrm{runs}}$. See Section~\ref{section:binary_sequence_metrics} for a definition of these metrics.}
\label{tab:design_ladder}
\begin{tabular}{l|c|cccc|ccccc}
\toprule
 & $\mathrm{BPB}$ & Engl & Germ & Code & Math & $\mathrm{B}$ & $C_{emp}$ & $H_g$ & $R_{\mathrm{CUSUM}}$ & $Z_{\mathrm{runs}}$ \\
\midrule
\textbf{Equal-Size} & 0.8893 & 1.007 & 1.017 & 0.5523 & 0.5961 & 1.001 & 4.995 & 0 & 1.575 & 124 \\
\midrule
\textbf{H-Net} & 0.6701 & 0.7778 & 0.7464 & 0.373 & 0.413 & 1.19 & 4.975 & 0.768 & 98.58 & 28.24 \\
\textbf{+Byte-Level Smooth} & 0.6571 & 0.7614 & 0.7321 & 0.3687 & 0.4112 & 2.612 & 4.986 & 0.74 & 76.77 & 63.07 \\
\textbf{-Confidence-Weight} & 0.6571 & 0.7614 & 0.7312 & 0.369 & 0.4102 & 2.81 & 4.968 & 0.7391 & 102.2 & 38.45 \\
\textbf{+Sigmoid Chunking} & 0.6578 & 0.7621 & 0.7331 & 0.3693 & 0.4125 & 2.606 & 4.986 & 0.7224 & 82.04 & 56.32 \\
\textbf{+CAB Loss (0.01)} & 0.6568 & 0.7608 & 0.7315 & 0.3696 & 0.4097 & 3.035 & 4.999 & 0.735 & 99.97 & 20.23 \\
\midrule
\textbf{H-Net+CAB Loss} & 0.664 & 0.7685 & 0.7384 & 0.3756 & 0.416 & 2.774 & 4.977 & 0.7421 & 102.8 & -4.4 \\
\midrule
\textbf{H-Net Scale Up} & 0.6126 & 0.7136 & 0.6759 & 0.3354 & 0.378 & 1.998 & 4.588 & 0.7533 & 83.69 & 20 \\
\textbf{\Sombrero Scale Up} & 0.6128 & 0.7139 & 0.6749 & 0.3364 & 0.376 & 3.133 & 4.604 & 0.7117 & 102.5 & 1.691 \\
\bottomrule
\end{tabular}
\end{table*}

\subsection{Setting}
One of the major challenges for learned sequence chunking is that natural text is intrinsically heterogeneous, interleaving multiple natural languages with formal sub-languages such as programming code, mathematical notation, and structured symbols, often within the same document or even the same sentence. We therefore curate a diverse data mix containing English and German natural language (future work should extend beyond two languages) as well as math and code. See Section~\ref{appendix:data} for details.

We FLOP-match all architectures to $1.95\pm0.01$ GFLOPs/byte. This matches the FLOPs/byte of our baseline architecture: a 1B-parameter Transformer with vocabulary size $151{,}936$ (average compression rate $4.35$ on our data mix). The baseline uses 13 Transformer layers with causal multi-head attention (13 heads, QK-norm), a SwiGLU MLP with multiplier $10/3$, and hidden size $13 \cdot 128 = 1664$.

We focus on HAT architectures with two stages and leave exploration of deeper nested architectures to future work. We largely follow H-Net in using Mamba-2 layers for encoder and decoder \citep{gu2023mamba,dao2024mamba2} and Transformer layers for the backbone \citep{vaswani2017attention}. Based on preliminary experiments, we decided to increase the number of Mamba-2 layers in encoder and decoder to 7, while keeping hidden dimensionality at 768, and setting $chunk\_size=256$, $d\_conv=4$, $d\_state=128$, and $expand=2$. FLOP-Matching for a target compression rate $C_{tar}=5.0$, we configure the backbone to consist of 16 Transformer layers with hidden dimensionality 2048, SwiGLU MLP expansion of $3.25$ \citep{shazeer2020glu} and causal multi-head attention with 16 heads and no QK-norm \citep{henry2020qknorm}. The resulting architecture has 0.98B parameters. 

We use a sequence length of $16{,}384$ bytes with document packing and a global batch size of 1024. We train for $50{,}000$ steps, resulting in a training budget of 839B bytes. We use AdamW \citep{loshchilov2019adamw} with gradient norm clipping at 1.0, and a warmup-stable-decay learning rate schedule \citep{wen2024wsd} with 500 warmup steps and 2,500 steps of square-root decay to 0. We use a reference learning rate of $\alpha_r=0.00125$ and a reference hidden dimensionality of $w_r=1536$. We set the actual learning rate of a layer with hidden dimensionality $w$ to $\alpha=\alpha_r \sqrt{w_r / w}$. H-Net uses a loss weight of $\omega=0.03$ for the ratio loss. We found that this weight results in a substantial mismatch of empirical and target compression rate ($C_{emp}=3.95$ versus $C_{tar}=5.0$), as also observed by \citet{hwang2025hnet}. Since such a mismatch ruins the FLOP-matching that was done for the target compression rate, we increase the ratio loss weight to $\omega=1.0$, which results in $C_{emp}=4.97$. 

\subsection{Results}
Our main results are summarized in Table~\ref{tab:design_ladder} (training curves of selected settings and metrics are shown in Figure~\ref{figure:training_curves} in the appendix). We report global bits-per-byte (BPB; lower is better) and BPB separately for the four domains English, German, Code, and Math. The overall BPB ranking transfers nearly perfectly to all sub-domains, indicating that our findings are generally applicable rather than domain-specific. We also report the average empirical compression $C_{emp}$, which is close to the target $C_{tar}=5.0$ for all settings so results are comparable. Moreover, we report our proposed boundary enrichment metric ($\mathrm{B}$; see Section~\ref{subsection:boundary_enrichment_metric}); the corresponding null-calibrated $Z_B$ scores (not shown in Table~\ref{tab:design_ladder}) are above $30$ for all configurations except for the equal-size baseline. We also report several binary sequence metrics (see Section~\ref{section:binary_sequence_metrics}). All values are averaged over $32{,}768$ (packed, unseen) validation sequences of length $16{,}384$ bytes.

We structure our experiments as a design ladder, starting from H-Net. H-Net substantially outperforms the equal-size split baseline (every fifth byte being a boundary) by over $0.20$ BPB, but its boundaries are not strongly aligned with hard predictions ($B$ remains close to the baseline value of $1$). We then add, modify, and remove components from H-Net. Replacing chunk-level with byte-level smoothing yields a substantial BPB improvement ($-0.013$) and markedly improves alignment of boundaries with hard predictions ($B=2.612$). With byte-level smoothing, we can also simplify other parts of the architecture: compared to H-Net, we replace confidence-weighted fusion of encoder and backbone by a simple residual addition of the two streams while keeping BPB unchanged. We further replace the cosine parametrization of the chunker by the conceptually simpler sigmoid parametrization; this incurs a minor BPB regression while still substantially outperforming H-Net.

Finally, adding the CAB loss with weight $0.01$ yields the final \Sombrero configuration. \Sombrero achieves the lowest BPB, the highest boundary enrichment score ($B=3.035$), and matches the target compression rate $C_{tar}$ closely. Adding CAB to H-Net directly improves BPB by $0.037$ and substantially increases boundary enrichment, consistent with CAB aligning boundaries with hard bytes.

\begin{figure}[tb]
\includegraphics[width=.95\linewidth]{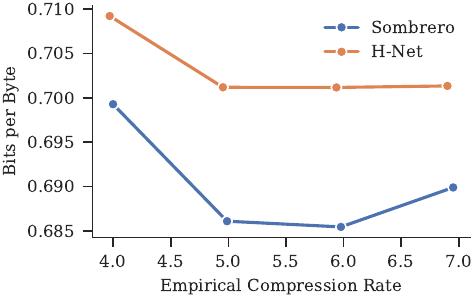}
\caption{Empirical compression $C_{emp}$ versus bits-per-byte when varying target compression $C_{tar}$ for 1B \Sombrero and H-Net.}
\label{figure:compression_sweep}
\end{figure}

\begin{table*}[tb]
\caption{Downstream evaluation results. ACC refers to accuracy of normalized loglikelihood, $C$ to empirical compression rate. ARC refers to AI2 ARC \citep{clark2018arc}, HSwag to HellaSwag \citep{zellers2019hellaswag}, MMLU to Measuring Massive Multitask Language Understanding \citep{hendrycks2020mmlu}, and TrQA to TruthfulQA \citep{lin2021truthfulqa}. DE refers to German versions of the evals.}
\label{tab:evals}
\begin{scriptsize}
{\setlength{\tabcolsep}{4pt}
\begin{tabular}{l|cc|cc|cc|cc|cc|cc|cc|cc|cc}
\toprule
& \multicolumn{2}{c}{ARC} & \multicolumn{2}{c}{ARC DE} & \multicolumn{2}{c}{HSwag} & \multicolumn{2}{c}{HSwag DE} & \multicolumn{2}{c}{MMLU} & \multicolumn{2}{c}{MMLU DE} & \multicolumn{2}{c}{TrQA} & \multicolumn{2}{c}{TrQA DE} & \multicolumn{2}{c}{Avg} \\
 & Acc & $C$ & Acc & $C$ & Acc & $C$ & Acc & $C$ & Acc & $C$ & Acc & $C$ & Acc & $C$ & Acc & $C$ & Acc & $C$ \\
\midrule
\textbf{Qwen Tokenizer} & 0.497 & 4.92 & 0.267 & 3.58 & 0.511 & 4.64 & 0.424 & 3.68 & 0.303 & 4.2 & 0.296 & 3.33 & 0.308 & 4 & 0.335 & 3.29 & 0.368 & 3.95 \\
\textbf{Custom Tokenizer} & 0.501 & 4.92 & 0.289 & 4.93 & 0.526 & 4.68 & 0.333 & 4.65 & 0.313 & 4.31 & 0.318 & 4.38 & 0.293 & 4.08 & 0.373 & 4.34 & 0.368 & 4.54 \\
\textbf{H-Net} & 0.509 & 5.44 & 0.303 & 5.38 & 0.563 & 4.8 & 0.465 & 4.78 & 0.324 & 4.87 & 0.326 & 4.91 & 0.346 & 4.4 & 0.376 & 4.55 & 0.402 & 4.89 \\
\textbf{\Sombrero} & 0.524 & 5.99 & 0.305 & 4.98 & 0.591 & 4.9 & 0.485 & 4.73 & 0.326 & 5.3 & 0.315 & 4.97 & 0.308 & 4.37 & 0.368 & 4.54 & 0.403 & 5.03 \\
\midrule
\textbf{Qwen Tokenizer 3B} & 0.589 & 4.92 & 0.358 & 3.58 & 0.634 & 4.64 & 0.424 & 3.68 & 0.403 & 4.2 & 0.367 & 3.33 & 0.29 & 4 & 0.376 & 3.29 & 0.43 & 3.97 \\
\textbf{Custom Tokenizer 3B} & 0.583 & 4.92 & 0.341 & 4.93 & 0.639 & 4.68 & 0.485 & 4.65 & 0.405 & 4.31 & 0.384 & 4.38 & 0.332 & 4.08 & 0.378 & 4.34 & 0.443 & 4.51 \\
\textbf{H-Net Scale Up} & 0.601 & 4.93 & 0.387 & 4.54 & 0.679 & 4.54 & 0.515 & 4.46 & 0.452 & 4.47 & 0.411 & 4.33 & 0.354 & 4.03 & 0.387 & 4.3 & 0.473 & 4.44 \\
\textbf{\Sombrero Scale Up} & 0.609 & 5.03 & 0.376 & 4.49 & 0.673 & 4.45 & 0.515 & 4.45 & 0.413 & 4.76 & 0.386 & 4.55 & 0.349 & 4.32 & 0.38 & 4.53 & 0.463 & 4.55 \\
\bottomrule
\end{tabular}
}
\end{scriptsize}
\end{table*}

\paragraph{Boundary Analysis.}
To characterize learned boundary sequences $b_{1:t}$ beyond the empirical compression rate $C_{emp}$ and boundary enrichment $B$, we report \emph{binary sequence metrics} that summarize spacing regularity, temporal stationarity, and short-range dependence. We measure (i) the normalized entropy of inter-boundary gaps $H_g$ (distances between consecutive $1$s), where lower values indicate more constrained/structured spacing, (ii) the CUSUM range $R_{\mathrm{CUSUM}}$ of the mean-centered cumulative sum, where larger values indicate stronger temporal variation in boundary rate (e.g., regime changes), and (iii) the Wald--Wolfowitz runs-test z-score $Z_{\mathrm{runs}}$, where values near $0$ are consistent with an i.i.d. Bernoulli baseline at the same marginal rate, and large positive values indicate fewer runs than expected (temporal persistence). Formal definitions are provided in Section~\ref{section:binary_sequence_metrics}. We provide qualitative illustrations of the learned sequence chunking in Section~\ref{subsection:chunk_illustration}.

Equal-Size by construction yields perfectly periodic spacing ($H_g=0$) and an almost constant boundary rate ($R_{\mathrm{CUSUM}}=1.6$), but extreme persistence ($Z_{\mathrm{runs}}=124$), consistent with rigid, non-adaptive segmentation. H-Net substantially improves BPB while producing diverse gap lengths ($H_g\approx0.77$), suggesting content-adaptive spacing, but exhibits strong temporal variation in boundary rate ($R_{\mathrm{CUSUM}}\approx99$) and non-trivial persistence ($Z_{\mathrm{runs}}\approx28$).

Across the \Sombrero ladder, these metrics indicate a shift in how stability is achieved. Byte-level smoothing improves BPB and reduces temporal rate variation ($R_{\mathrm{CUSUM}}: 99\!\rightarrow\!77$) but increases persistence ($Z_{\mathrm{runs}}: 28\!\rightarrow\!63$). Removing confidence weighting leaves BPB unchanged while worsening stationarity ($R_{\mathrm{CUSUM}}\approx102$) and reducing persistence ($Z_{\mathrm{runs}}\approx38$), suggesting confidence weighting traded local volatility for inertia. Sigmoid chunking further constrains spacing ($H_g\approx0.72$) while remaining relatively persistent ($Z_{\mathrm{runs}}\approx56$). Finally, adding the CAB loss (final \Sombrero) yields the best BPB and the lowest persistence among learned methods ($Z_{\mathrm{runs}}\approx20$) while maintaining the target compression ($C_{emp}\approx5$), indicating more locally responsive boundary decisions even though global rate variation remains ($R_{\mathrm{CUSUM}}\approx100$).

\paragraph{Varying the target compression rate.}
We conduct an additional experiment in which we vary the target compression rate, $C_{tar} \in \{4, 5, 6, 7\}$, for H-Net and \Sombrero, while keeping each setting FLOP-matched to $1.95$ GFLOPs/byte. This results in models with $0.66$B, $0.98$B, $1.22$B, and $1.52$B parameters, respectively. For computational reasons, we train only for $20{,}000$ steps in this experiment. Results are shown in Figure~\ref{figure:compression_sweep}. We observe that in all cases, the empirical compression rate closely matches the target compression rate. For every compression rate, \Sombrero substantially outperforms H-Net. Moreover, both \Sombrero and H-Net benefit from increasing the compression rate beyond the $4.6$ achieved by a tokenizer on average for our data.

\paragraph{Scaling Up.}
We scale up H-Net and \Sombrero to $5.85$ GFLOPs/byte and $2.19$B parameters by increasing encoder and decoder to 10 Mamba-2 layers with hidden size 1280. Moreover, we increase the backbone to 21 Transformer layers with 21 heads and hidden size 2688. We set the target compression rate to $C_{tar}=4.6$ to match the compression rate achieved by a tokenizer fitted on this domain. We also add QK-norm to prevent potential instabilities from scaling up (though we did not observe any).

Somewhat surprisingly, H-Net and \Sombrero perform essentially on par in terms of BPB at this scale, even though \Sombrero learns much more hardness-aligned boundaries ($B=3.1$) compared to H-Net ($B=2.0$). Notably, H-Net benefits from scale in terms of boundary enrichment ($B=2.0$ at scale-up versus $B=1.2$ at smaller scale). We suspect that boundary quality may saturate earlier than perplexity, and that stronger encoders and decoders make the backbone less of a bottleneck, reducing the importance of hardness-aligned chunking. Accordingly, we consider a better understanding of the scaling behavior of hierarchical sequence models a core future research direction.

\paragraph{Downstream Evaluations}
We evaluate H-Net and \Sombrero on several downstream log-likelihood-based evaluations and compare them to FLOP-matched subword tokenizer baselines: a ``custom'' tokenizer fitted on our data mix and the Qwen3 tokenizer \citep{yang2025qwen3technicalreport}. We compare at both $1.95$ and $5.85$ GFLOPs/Byte scales, referring to the former as ``default'' and latter as the ``scale-up'' regime. Results are summarized in Table \ref{tab:evals}. Overall, H-Net and \Sombrero outperform both tokenizer baselines on both scales consistently and hit the respective target compression rates closely. \Sombrero obtains higher average accuracy at higher compression than H-Net on the default scale; both are roughly on par in the scale-up regime (slightly lower average accuracy at higher compression for \Sombrero). 

\section{Conclusion}
We revisited hierarchical sequence models through the lens of \emph{adaptive compute allocation}, with the central hypothesis that learned chunk boundaries should concentrate backbone compute on positions that are hard to predict (high next-byte surprisal). To make this hypothesis testable, we introduced \emph{boundary enrichment} as a router-agnostic, quantitative measure of boundary quality, and we proposed the \emph{confidence-alignment boundary} loss to directly encourage hardness-aligned boundary placement during end-to-end training. Finally, we simplified and stabilized learned chunking by moving confidence-weighted smoothing to the byte level, which improves the accuracy--efficiency trade-off at 1B scale without materially increasing compute.

Taken together, these contributions strengthen the case for hierarchical sequence models as a practical alternative to fixed subword tokenizers, while also providing diagnostics and levers to reason about (and steer) where expensive backbone compute is spent. Directions for future work include understanding scaling behavior across model size and training duration, extending the analysis to deeper hierarchies and broader language/domain coverage, and improving training and inference efficiency (e.g., better batching and hardware-aware implementations for dynamic compression).

\section*{Impact Statement}
This paper presents work whose goal is to advance the field of machine learning. There are many potential societal consequences of our work, none of which we feel must be specifically highlighted here.

\bibliography{references}

@article{gage1994bpe,
  title={A New Algorithm for Data Compression},
  author={Gage, Philip},
  journal={The C Users Journal},
  year={1994}
}

@article{radford2019gpt2,
  title={Language Models are Unsupervised Multitask Learners},
  author={Radford, Alec and Wu, Jeffrey and Child, Rewon and Luan, David and Amodei, Dario and Sutskever, Ilya},
  year={2019},
  journal = {OpenAI},
  note={OpenAI technical report}
}

@article{xue2022byt5,
  title={ByT5: Towards a Token-Free Future with Pre-trained Byte-to-Byte Models},
  author={Xue, Linting and Barua, Aditya and Constant, Noah and Al-Rfou, Rami and Narang, Sharan and Kale, Mihir and Roberts, Adam and Raffel, Colin},
  journal={Transactions of the Association for Computational Linguistics},
  year={2022},
  doi={10.1162/tacl_a_00466},
  eprint={2105.13626},
  archivePrefix={arXiv},
  primaryClass={cs.CL}
}

@article{clark2022canine,
  title={CANINE: Pre-training an Efficient Tokenization-Free Encoder for Language Representation},
  author={Clark, Jonathan H. and Garrette, Dan and Turc, Iulia and Wieting, John},
  journal={Transactions of the Association for Computational Linguistics},
  year={2022},
  doi={10.1162/tacl_a_00448},
  eprint={2103.06874},
  archivePrefix={arXiv},
  primaryClass={cs.CL}
}

@inproceedings{pruthi2019misspellings,
  title={Combating Adversarial Misspellings with Robust Word Recognition},
  author={Pruthi, Danish and Dhingra, Bhuwan and Lipton, Zachary C.},
  booktitle={Proceedings of the 57th Annual Meeting of the Association for Computational Linguistics},
  year={2019},
  pages={5582--5591},
  doi={10.18653/v1/P19-1561}
}

@article{geh2025adversarialtokenization,
  title={Adversarial Tokenization},
  author={Geh, Ryan L. and others},
  year={2025},
  journal = {arXiv e-prints},
  eprint={2503.02174},
  archivePrefix={arXiv},
  primaryClass={cs.CL}
}

@inproceedings{sennrich2016bpe,
  title={Neural Machine Translation of Rare Words with Subword Units},
  author={Sennrich, Rico and Haddow, Barry and Birch, Alexandra},
  booktitle={Proceedings of the 54th Annual Meeting of the Association for Computational Linguistics (Volume 1: Long Papers)},
  year={2016},
  pages={1715--1725},
  doi={10.18653/v1/P16-1162},
  url={https://aclanthology.org/P16-1162/}
}

@inproceedings{kudo2018sentencepiece,
  title={SentencePiece: A simple and language independent subword tokenizer and detokenizer for Neural Text Processing},
  author={Kudo, Taku and Richardson, John},
  booktitle={Proceedings of the 2018 Conference on Empirical Methods in Natural Language Processing: System Demonstrations},
  year={2018},
  pages={66--71},
  url={https://aclanthology.org/D18-2012/},
  eprint={1808.06226},
  archivePrefix={arXiv},
  primaryClass={cs.CL}
}

@inproceedings{schuster2012wordpiece,
  title={Japanese and Korean Voice Search},
  author={Schuster, Mike and Nakajima, Kaisuke},
  booktitle={2012 IEEE International Conference on Acoustics, Speech and Signal Processing (ICASSP)},
  year={2012},
  pages={5149--5152},
  doi={10.1109/ICASSP.2012.6289079},
  url={https://research.google/pubs/japanese-and-korean-voice-search/}
}

@inproceedings{tay2021charformer,
  title={{Charformer}: Fast Character Transformers via Gradient-based Subword Tokenization},
  author={Tay, Yi and Tran, Vinh Q. and Ruder, Sebastian and Gupta, Jai and Chung, Hyung Won and Bahri, Dara and Qin, Zhen and Baumgartner, Simon and Yu, Cong and Metzler, Donald},
  booktitle={International Conference on Learning Representations (ICLR)},
  year={2022},
  url={https://openreview.net/forum?id=JtBRnrlOEFN}
}

@inproceedings{elb2020characterbert,
  title={{CharacterBERT}: Reconciling {ELMo} and {BERT} for Word-Level Open-Vocabulary Representations From Characters},
  author={El Boukkouri, Hicham and Ferret, Olivier and Lavergne, Thomas and Noji, Hiroshi and Zweigenbaum, Pierre and Tsujii, Jun'ichi},
  booktitle={Proceedings of the 28th International Conference on Computational Linguistics (COLING)},
  year={2020},
  url={https://aclanthology.org/2020.coling-main.609/}
}

@inproceedings{yu2023megabyte,
  title={{MEGABYTE}: Predicting Million-byte Sequences with Multiscale Transformers},
  author={Yu, Lili and Simig, D{\'a}niel and Flaherty, Colin and Aghajanyan, Armen and Zettlemoyer, Luke and Lewis, Mike},
  booktitle={International Conference on Machine Learning (ICML)},
  year={2023},
  pages={8535--8558}
}

@inproceedings{wang2024mambabyte,
  title={{MambaByte}: Token-free Selective State Space Model},
  author={Wang, Junxiong and Gangavarapu, Tushaar and Yan, Jing Nathan and Rush, Alexander M.},
  booktitle={Conference on Language Modeling (COLM)},
  year={2024},
  url={https://openreview.net/pdf?id=X1xNsuKssb}
}

@inproceedings{slagle2024spacebyte,
  title={{SpaceByte}: Towards Deleting Tokenization from Large Language Modeling},
  author={Slagle, Kevin},
  booktitle={Advances in Neural Information Processing Systems (NeurIPS)},
  year={2024},
}

@inproceedings{kallini2024mrt5,
  title={{MrT5}: Dynamic Token Merging for Efficient Byte-level Language Models},
  author={Kallini, Julie and Murty, Shikhar and Manning, Christopher D. and Potts, Christopher and Csord{\'a}s, R{\'o}bert},
  booktitle={International Conference on Learning Representations (ICLR)},
  year={2025},
  url={https://iclr.cc/virtual/2025/poster/29408}
}

@inproceedings{pagnoni2024blt,
  title={{Byte Latent Transformer}: Patches Scale Better Than Tokens},
  author={Pagnoni, Artidoro and Pasunuru, Ram and Rodriguez, Pedro and Nguyen, John and Muller, Benjamin and Li, Margaret and Zhou, Chunting and Yu, Lili and Weston, Jason and Zettlemoyer, Luke and Ghosh, Gargi and Lewis, Mike and Holtzman, Ari and Iyer, Srinivasan},
  booktitle={Proceedings of the 63rd Annual Meeting of the Association for Computational Linguistics (Volume 1: Long Papers)},
  year={2025},
  url={https://aclanthology.org/2025.acl-long.453/}
}

@misc{neitemeier2025hat,
  title={Hierarchical Autoregressive Transformers: Combining Byte- and Word-Level Processing for Robust, Adaptable Language Models},
  author={Neitemeier, Pit and Deiseroth, Bj{\"o}rn and Eichenberg, Constantin and Balles, Lukas},
  year={2025},
  eprint={2501.10322},
  archivePrefix={arXiv},
  primaryClass={cs.CL},
  url={https://arxiv.org/abs/2501.10322}
}

@misc{egli2025mblm,
  title={Multiscale Byte Language Models -- A Hierarchical Architecture for Causal Million-Length Sequence Modeling},
  author={Egli, Eric and Manica, Matteo and Born, Jannis},
  year={2025},
  eprint={2502.14553},
  archivePrefix={arXiv},
  primaryClass={cs.CL},
  url={https://arxiv.org/abs/2502.14553}
}

@misc{videau2025aunet,
  title={From Bytes to Ideas: Language Modeling with Autoregressive U-Nets},
  author={Videau, Mathurin and Idrissi, Badr Youbi and Leite, Alessandro and Schoenauer, Marc and Teytaud, Olivier and Lopez-Paz, David},
  year={2025},
  eprint={2506.14761},
  archivePrefix={arXiv},
  primaryClass={cs.CL},
  url={https://arxiv.org/abs/2506.14761}
}

@misc{hwang2025hnet,
  title={Dynamic Chunking for End-to-End Hierarchical Sequence Modeling},
  author={Hwang, Sukjun and Wang, Brandon and Gu, Albert},
  year={2025},
  eprint={2507.07955},
  archivePrefix={arXiv},
  primaryClass={cs.CL},
  url={https://arxiv.org/abs/2507.07955}
}

@misc{zakershahrak2025hnetpp,
  title={H-Net++: Hierarchical Dynamic Chunking for Tokenizer-Free Language Modelling in Morphologically-Rich Languages},
  author={Zakershahrak, Mehrdad and Ghodratnama, Samira},
  year={2025},
  eprint={2508.05628},
  archivePrefix={arXiv},
  primaryClass={cs.CL},
  url={https://arxiv.org/abs/2508.05628}
}

@misc{minixhofer2025bolmo,
  title={Bolmo: Byteifying the Next Generation of Language Models},
  author={Minixhofer, Benjamin and Murray, Tyler and Limisiewicz, Tomasz and Korhonen, Anna and Zettlemoyer, Luke and Smith, Noah A. and Ponti, Edoardo M. and Soldaini, Luca and Hofmann, Valentin},
  year={2025},
  eprint={2512.15586},
  archivePrefix={arXiv},
  primaryClass={cs.CL},
  url={https://arxiv.org/abs/2512.15586}
}

@misc{li2025bytegen,
  title={ByteGen: A Tokenizer-Free Generative Model for Orderbook Events in Byte Space},
  author={Li, Yang and Chen, Zhi},
  year={2025},
  eprint={2508.02247},
  archivePrefix={arXiv},
  primaryClass={cs.LG},
  url={https://arxiv.org/abs/2508.02247}
}

@misc{evabyte2025,
  title={EvaByte: Efficient Byte-level Language Models at Scale},
  author={Zheng, Lin and Zhao, Xueliang and Wang, Guangtao and Wu, Chen and Dong, David and Wang, Angela and Wang, Mingran and Du, Yun and Bo, Haige and Sharma, Amol and Li, Bo and Zhang, Kejie and Hu, Changran and Thakker, Urmish and Kong, Lingpeng},
  year={2025},
  howpublished={Technical report / model release},
  url={https://hkunlp.github.io/blog/2025/evabyte/}
}

@ARTICLE{2013arXiv1305.2982B,
       author = {{Bengio}, Yoshua},
        title = "{Estimating or Propagating Gradients Through Stochastic Neurons}",
      journal = {arXiv e-prints},
         year = 2013,
        month = may,
          eid = {arXiv:1305.2982},
        pages = {arXiv:1305.2982},
          doi = {10.48550/arXiv.1305.2982},
archivePrefix = {arXiv},
       eprint = {1305.2982},
 primaryClass = {cs.LG},
       adsurl = {https://ui.adsabs.harvard.edu/abs/2013arXiv1305.2982B}
}

@inproceedings{su2024nemotroncc,
  title={Nemotron-{CC}: Transforming Common Crawl into a Refined Long-Horizon Pretraining Dataset},
  author={Su, Dan and Kong, Kezhi and Lin, Ying and Jennings, Joseph and Norick, Brandon and Kliegl, Markus and Patwary, Mostofa and Shoeybi, Mohammad and Catanzaro, Bryan},
  booktitle={Proceedings of the 63rd Annual Meeting of the Association for Computational Linguistics (Volume 1: Long Papers)},
  year={2025},
  pages={2459--2475},
  url={https://aclanthology.org/2025.acl-long.123/}
}

@article{penedo2025fineweb2,
  title={FineWeb2: One Pipeline to Scale Them All},
  author={Penedo, Guilherme and others},
  journal={arXiv preprint arXiv:2506.20920},
  year={2025}
}

@inproceedings{cohan2018discourse,
  title={A Discourse-Aware Attention Model for Abstractive Summarization of Long Documents},
  author={Cohan, Arman and Dernoncourt, Franck and Kim, Doo Soon and Bui, Trung and Kim, Seokhwan and Chang, Walter and Goharian, Nazli},
  booktitle={NAACL},
  year={2018}
}

@inproceedings{banon2020paracrawl,
  title={ParaCrawl: Web-Scale Acquisition of Parallel Corpora},
  author={Ba{\~n}{\'o}n, Marta and Chen, Pinzhen and Haddow, Barry and Heafield, Kenneth and Hoang, Hieu and Koehn, Philipp and Liu, Qun and Miceli Barone, Antonio Valerio and Sennrich, Rico and Thompson, Brian},
  booktitle={ACL},
  year={2020}
}

@inproceedings{saxton2019mathematicsdataset,
  title={Analysing Mathematical Reasoning Abilities of Neural Models},
  author={Saxton, David and Grefenstette, Edward and Hill, Felix and Kohli, Pushmeet},
  booktitle={ICLR},
  year={2019}
}

@article{mitra2024orcamath,
  title={Orca-Math: Unlocking the potential of SLMs in Grade School Math},
  author={Mitra, Arindam and Khanpour, Hamed and Rosset, Corby and Awadallah, Ahmed},
  journal={arXiv preprint arXiv:2402.14830},
  year={2024}
}

@article{kocetkov2022stack,
  title={The Stack: 3 {TB} of Permissively Licensed Source Code},
  author={Kocetkov, Denis and Li, Raymond and Ben Allal, Loubna and Li, Jia and Mou, Chenghao and Jernite, Yacine and Mitchell, Margaret and Mu{\~n}oz Ferrandis, Carlos and Hughes, Sean and Wolf, Thomas and Bahdanau, Dzmitry and Von Werra, Leandro and de Vries, Harm},
  journal={Transactions on Machine Learning Research},
  year={2023},
  url={https://openreview.net/forum?id=pxpbTdUEpD},
  note={Published 07 Jun 2023; accepted by TMLR}
}

@article{lozhkov2024starcoder2,
  title={StarCoder2 and The Stack v2: The Next Generation},
  author={Lozhkov, Anton and others},
  journal={arXiv preprint arXiv:2402.19173},
  year={2024}
}

@article{li2022alphacode,
  title={Competition-Level Code Generation with AlphaCode},
  author={Li, Yujia and others},
  journal={Science},
  year={2022}
}

@inproceedings{hendrycks2021apps,
  title={Measuring Coding Challenge Competence With {APPS}},
  author={Hendrycks, Dan and Basart, Steven and Kadavath, Saurav and Mazeika, Mantas and Arora, Akul and Guo, Ethan and Burns, Collin and Puranik, Samir and He, Horace and Song, Dawn and Steinhardt, Jacob},
  booktitle={Proceedings of the Neural Information Processing Systems Track on Datasets and Benchmarks (NeurIPS Datasets and Benchmarks)},
  year={2021},
  url={https://datasets-benchmarks-proceedings.neurips.cc/paper/2021/hash/c24cd76e1ce41366a4bbe8a49b02a028-Abstract-round2.html}
}

@inproceedings{nanz2014rosettacode,
  title={A Comparative Study of Programming Languages in {Rosetta} Code},
  author={Nanz, Sebastian and Furia, Carlo A.},
  booktitle={Proceedings of the 37th International Conference on Software Engineering (ICSE)},
  year={2015},
  doi={10.5555/2818754.2818848}
}

@article{burns2025alephalphagermanweb,
  title={Aleph-Alpha-GermanWeb: Improving German-language LLM pre-training with model-based data curation and synthetic data generation},
  author={Burns, Thomas F and Parcalabescu, Letitia and W{\"a}ldchen, Stephan and Barlow, Michael and Ziegltrum, Gregor and Stampa, Volker and Harren, Bastian and Deiseroth, Bj{\"o}rn},
  journal={arXiv preprint arXiv:2505.00022},
  year={2025},
  doi={10.48550/arXiv.2505.00022}
}

@misc{yang2025qwen3technicalreport,
      title={Qwen3 Technical Report}, 
      author={An Yang and Anfeng Li and Baosong Yang and Beichen Zhang and Binyuan Hui and Bo Zheng and Bowen Yu and Chang Gao and Chengen Huang and Chenxu Lv and Chujie Zheng and Dayiheng Liu and Fan Zhou and Fei Huang and Feng Hu and Hao Ge and Haoran Wei and Huan Lin and Jialong Tang and Jian Yang and Jianhong Tu and Jianwei Zhang and Jianxin Yang and Jiaxi Yang and Jing Zhou and Jingren Zhou and Junyang Lin and Kai Dang and Keqin Bao and Kexin Yang and Le Yu and Lianghao Deng and Mei Li and Mingfeng Xue and Mingze Li and Pei Zhang and Peng Wang and Qin Zhu and Rui Men and Ruize Gao and Shixuan Liu and Shuang Luo and Tianhao Li and Tianyi Tang and Wenbiao Yin and Xingzhang Ren and Xinyu Wang and Xinyu Zhang and Xuancheng Ren and Yang Fan and Yang Su and Yichang Zhang and Yinger Zhang and Yu Wan and Yuqiong Liu and Zekun Wang and Zeyu Cui and Zhenru Zhang and Zhipeng Zhou and Zihan Qiu},
      year={2025},
      eprint={2505.09388},
      archivePrefix={arXiv},
      primaryClass={cs.CL},
      url={https://arxiv.org/abs/2505.09388}, 
}

@misc{owodunni2025flexitokens,
  title={FLEXITOKENS: Flexible Tokenization for Evolving Language Models},
  author={Owodunni, Abraham Toluwase and Ahia, Orevaoghene and Kumar, Sachin},
  year={2025},
  eprint={2507.12720},
  archivePrefix={arXiv},
  primaryClass={cs.CL},
  doi={10.48550/arXiv.2507.12720},
  url={https://arxiv.org/abs/2507.12720}
}

@misc{liu2025superbpe,
  title={SuperBPE: Space Travel for Language Models},
  author={Liu, Alisa and Hayase, Jonathan and Hofmann, Valentin and Oh, Sewoong and Smith, Noah A. and Choi, Yejin},
  year={2025},
  eprint={2503.13423},
  archivePrefix={arXiv},
  primaryClass={cs.CL},
  doi={10.48550/arXiv.2503.13423},
  url={https://arxiv.org/abs/2503.13423}
}

@misc{dolga2025hierarchicalbpe,
  title={Dynamic Grouping with Hierarchical {BPE}},
  author={Dolga, L. and others},
  year={2025},
  eprint={2510.15517},
  archivePrefix={arXiv},
  primaryClass={cs.CL},
  doi={10.48550/arXiv.2510.15517},
  url={https://arxiv.org/abs/2510.15517}
}

@inproceedings{vaswani2017attention,
  title={Attention Is All You Need},
  author={Vaswani, Ashish and Shazeer, Noam and Parmar, Niki and Uszkoreit, Jakob and Jones, Llion and Gomez, Aidan N. and Kaiser, Lukasz and Polosukhin, Illia},
  booktitle={Advances in Neural Information Processing Systems},
  year={2017}
}

@misc{gu2023mamba,
  title={Mamba: Linear-Time Sequence Modeling with Selective State Spaces},
  author={Gu, Albert and Dao, Tri},
  year={2023},
  eprint={2312.00752},
  archivePrefix={arXiv},
  primaryClass={cs.LG},
  doi={10.48550/arXiv.2312.00752}
}

@misc{dao2024mamba2,
  title={Mamba-2: Transformers are SSMs},
  author={Dao, Tri and Gu, Albert},
  year={2024},
  eprint={2405.21060},
  archivePrefix={arXiv},
  primaryClass={cs.LG},
  doi={10.48550/arXiv.2405.21060}
}

@inproceedings{shazeer2020glu,
  title={{GLU} Variants Improve Transformer},
  author={Shazeer, Noam},
  booktitle={arXiv e-prints},
  year={2020},
  eprint={2002.05202},
  archivePrefix={arXiv},
  primaryClass={cs.LG},
  doi={10.48550/arXiv.2002.05202}
}

@inproceedings{loshchilov2019adamw,
  title={Decoupled Weight Decay Regularization},
  author={Loshchilov, Ilya and Hutter, Frank},
  booktitle={International Conference on Learning Representations (ICLR)},
  year={2019}
}

@misc{henry2020qknorm,
  title={Query-Key Normalization for Transformers},
  author={Henry, Alex and Dachapally, Prudhvi and Pawar, Aniket and Chen, Yiren and Glass, James},
  year={2020},
  eprint={2010.04245},
  archivePrefix={arXiv},
  primaryClass={cs.LG},
  doi={10.48550/arXiv.2010.04245}
}

@misc{clark2018arc,
  title={Think you have Solved Question Answering? Try {ARC}, the {AI2} Reasoning Challenge},
  author={Clark, Peter and Cowhey, Isaac and Etzioni, Oren and Khot, Tushar and Sabharwal, Ashish and Schoenick, Carissa and Tafjord, Oyvind},
  year={2018},
  eprint={1803.05457},
  archivePrefix={arXiv},
  primaryClass={cs.AI},
  doi={10.48550/arXiv.1803.05457}
}

@misc{zellers2019hellaswag,
  title={HellaSwag: Can a Machine Really Finish Your Sentence?},
  author={Zellers, Rowan and Holtzman, Ari and Bisk, Yonatan and Farhadi, Ali and Choi, Yejin},
  year={2019},
  eprint={1905.07830},
  archivePrefix={arXiv},
  primaryClass={cs.CL},
  doi={10.48550/arXiv.1905.07830}
}

@misc{hendrycks2020mmlu,
  title={Measuring Massive Multitask Language Understanding},
  author={Hendrycks, Dan and Burns, Collin and Basart, Steven and Zou, Andy and Mazeika, Mantas and Song, Dawn and Steinhardt, Jacob},
  year={2020},
  eprint={2009.03300},
  archivePrefix={arXiv},
  primaryClass={cs.CY},
  doi={10.48550/arXiv.2009.03300}
}

@misc{lin2021truthfulqa,
  title={TruthfulQA: Measuring How Models Mimic Human Falsehoods},
  author={Lin, Stephanie and Hilton, Jacob and Evans, Owain},
  year={2021},
  eprint={2109.07958},
  archivePrefix={arXiv},
  primaryClass={cs.CL},
  doi={10.48550/arXiv.2109.07958}
}

@misc{wen2024wsd,
  title={Understanding Warmup-Stable-Decay Learning Rates: A River Valley Loss Landscape Perspective},
  author={Wen, Kaiyue and Li, Zhiyuan and Wang, Jason and Hall, David and Liang, Percy and Ma, Tengyu},
  year={2024},
  eprint={2410.05192},
  archivePrefix={arXiv},
  primaryClass={cs.LG},
  doi={10.48550/arXiv.2410.05192}
}
\bibliographystyle{icml2026}

\newpage
\appendix
\onecolumn
\section{Synthetic Problem} \label{appendix:synthetic_problem}
\paragraph{Synthetic piecewise-constant latent dynamics.}
We generate long sequences with latent change points and noisy linear observations to study boundary detection and compression. A sample consists of a length-$T$ sequence of latent states $\{z_t\}_{t=1}^T$ with $z_t \in \mathbb{R}^{d_z}$, observed states $\{x_t\}_{t=1}^T$ with $x_t \in \mathbb{R}^{d_x}$, and a binary boundary mask $\{b_t\}_{t=1}^T$ indicating change points. We draw boundaries independently as
\begin{equation}
b_t \sim \mathrm{Bernoulli}(p), \qquad b_1 = 1,
\end{equation}
where $p \in (0,1)$ controls the expected segment length (and thus the target compression rate $C_{tar}$) via
\begin{equation}
\mathbb{E}[\text{segment length}] = \frac{1}{p} = C_{tar}
\end{equation}
Let $1=\tau_1 < \tau_2 < \cdots < \tau_K$ be the indices where $b_{\tau_k}=1$ (change points), and define segments $[\tau_k, \tau_{k+1}-1]$ with $\tau_{K+1}=T+1$. For each segment $k$, we sample a segment-level latent prototype
\begin{equation}
\tilde z_k \sim \mathcal{N}(0, I_{d_z}),
\end{equation}
and set the latent sequence to be piecewise constant:
\begin{equation}
z_t = \tilde z_k \quad \text{for } t \in [\tau_k, \tau_{k+1}-1].
\end{equation}
Observations are generated by a fixed random linear map $W \in \mathbb{R}^{d_z \times d_x}$ (sampled once per generator instance) plus i.i.d.\ Gaussian noise:
\begin{equation}
x_t = z_t^\top W + \varepsilon_t, \qquad \varepsilon_t \sim \mathcal{N}(0, \sigma^2 I_{d_x}),
\end{equation}
with noise scale $\sigma$ controlling the signal-to-noise ratio. In all experiments we use $T=2^{15}$, $d_z=256$, and $d_x=32$ unless stated otherwise. This construction yields long-range sequences with abrupt regime switches, where the optimal representation is to compress each constant segment into a single latent prototype, and the expected compression target is set by $1/p$.

\paragraph{Model for sequence segmentation.}
We instantiate a minimal hierarchical segmentation model based on using three components from HAT (see Section \ref{subsection:background}): the \emph{encoder} $\mathcal{E}$ that produces contextual per-position features, the \emph{chunker} $\mathcal{C}_c$ that produces boundary scores $p$, boundary indicators $b$ and confidences $c$, and the \emph{temporal expansion} module that expands chunk-level activations back to a length-$T$ sequence. Given an observed sequence $x_{1:T}\in\mathbb{R}^{d_x}$, the encoder produces
\begin{equation}
h_{1:T} = \mathcal{E}(x_{1:T}) \in \mathbb{R}^{T\times d_h},
\end{equation}
where $\mathcal{E}$ is a Mamba-2 block operating on the full sequence. The backbone connector then predicts a binary boundary mask $\hat b_{1:T}\in\{0,1\}^T$, along with per-position boundary scores $\hat p_{1:T}\in(0,1)^T$ and associated confidences $\hat c_{1:T}\in[0,1]^T$:
\begin{equation}
(\hat  b_{1:T}, \hat  p_{1:T}, \hat  c_{1:T}) = \mathcal{C}_c(h_{1:T};C_{tar}),
\end{equation}
where $C_{tar}$ denotes the target compression factor. We consider two chunkers (see Section \ref{subsection:chunker}): (i) a cosine  chunker that scores change points based on cosine dissimilarity between adjacent (projected) encoder features, and (ii) a sigmoid chunker that maps encoder features to scores via a linear projection and logistic function.

\paragraph{Backbone connector and chunk-level activations.}
In the full model, chunk-level activations are obtained by running a backbone on a compressed sequence and using a backbone connector. In this synthetic setup, we replace the backbone by the ground-truth latent process: we read out the chunk-level states by sub-sampling the latent sequence at predicted boundaries,
\begin{equation}
\tilde z_{1:K} = \{ z_t \,:\, \hat b_t = 1 \}, \qquad K = \sum_{t=1}^T \hat b_t,
\end{equation}
which isolates the learning problem to segmentation (boundary placement) and temporal expansion, without introducing decoder or backbone modeling error.

\paragraph{Temporal expansion.}
The temporal expansion module $f_{\mathrm{exp}}$ maps chunk-level states back to a length-$T$ sequence using the predicted segmentation and confidence signals:
\begin{equation}
\hat z_{1:T} = f_{\mathrm{exp}}\!\left(\tilde z_{1:K}; p_{1:T}, c_{1:T}, b_{1:T}\right),
\end{equation}
where $f_{\mathrm{exp}}$ is either \emph{chunk-confidence-weighted} (operating at chunk granularity) or \emph{byte-confidence-weighted} (operating at position granularity), see Section \ref{subsection:byte_smoothing}. Both variants interpolate between repeating chunk states within a segment and smoothing across predicted boundaries based on the confidence signal (higher confidence yields sharper boundaries; lower confidence yields stronger mixing across adjacent chunks).

\paragraph{Training objective.}
We train end-to-end to reconstruct the ground-truth latent sequence $z_{1:T}$ (available in the synthetic generator) from the expanded predictions $\hat z_{1:T}$. The main term is a mean-squared error,
\begin{equation}
\mathcal{L}_{\mathrm{mse}} = \frac{1}{T}\sum_{t=1}^T \left\| \hat z_t - z_t \right\|_2^2,
\end{equation}
which encourages the model to place boundaries such that piecewise-constant latent regimes are preserved after compression and expansion. To control the achieved compression rate, we add a ratio loss that penalizes deviations of the predicted boundary process from a target compression $C_{tar}$:
\begin{equation}
\mathcal{L}_{\mathrm{ratio}} =
\mathrm{RatioLoss}\!\left(p_{1:T}, b_{1:T}; C_{tar}\right),
\qquad
\mathcal{L} = \mathcal{L}_{\mathrm{mse}} + \mathcal{L}_{\mathrm{ratio}}.
\end{equation}
The ratio loss term uses both the soft boundary scores and the realized mask to stabilize learning and to keep the expected number of boundaries close to $T/C_{tar}$, see \citet{hwang2025hnet}.

\paragraph{Optimization and variants.}
We optimize all parameters of the encoder and chunker with AdamW (learning rate $10^{-3}$) for $1,500$ steps. For each step, we draw a fresh synthetic sequence and perform a forward pass to obtain $(\hat z_{1:T}, \hat b_{1:T}, \hat p_{1:T}, \hat c_{1:T})$, compute $\mathcal{L}$, and update parameters via backpropagation. We evaluate four configurations formed by combining (i) cosine-similarity vs.\ sigmoid chunker and (ii) chunk-confidence-weighted vs.\ byte-confidence-weighted temporal expansion, keeping the target compression fixed (e.g., $C_{tar}=4$) and repeating experiments $25$ times across different random seeds. We report $\mathcal{L}_{\mathrm{mse}}$, $\mathcal{L}_{\mathrm{ratio}}$, $C_{emp}$, and the accuracy between predicted and ground-truth segment boundaries. Results are shown in Figure \ref{figure:synthetic_problem} and discussed in the main document.

\section{Data Mix} \label{appendix:data}
We construct our German web pretraining mixture using curated Common Crawl–style sources and model-based quality filtering, including Aleph-Alpha-GermanWeb for German web data curation and synthetic augmentation \citep{burns2025alephalphagermanweb}. For broad web-scale English and multilingual web data, we rely on modern Common Crawl refinement pipelines such as Nemotron-CC \citep{su2024nemotroncc} and FineWeb2 \citep{penedo2025fineweb2}. For scientific and technical text, we include arXiv-style paper corpora \citep{cohan2018discourse}. To support translation and bilingual alignment, we add ParaCrawl EN–DE \citep{banon2020paracrawl}. For reasoning and STEM, we incorporate the DeepMind Mathematics dataset and related math-focused resources \citep{saxton2019mathematicsdataset,mitra2024orcamath}. For code modeling, we include large permissively-licensed code corpora and derivatives (The Stack / StarCoder) \citep{kocetkov2022stack,lozhkov2024starcoder2}, alongside problem-solving benchmarks such as CodeContests/AlphaCode and APPS \citep{li2022alphacode,hendrycks2021apps}.

More specifically, we trained on the following data mix with the respective blending ratios: 

\begin{enumerate}[noitemsep,topsep=0pt]
  \item Nemotron-CC (high): 35.0
  \item Nemotron-CC (medium-high): 6.0
  \item Nemotron-CC (medium): 24.0
  \item High-Quality-All EN v3: 4.5
  \item arXiv: 0.5
  \item FineWeb2-DE Quality Classified v4 (high): 0.43
  \item FineWeb2-DE Quality Classified v4 (medium-high): 0.48
  \item FineWeb2-DE Quality Classified v4 (medium): 0.70
  \item FineWeb2-DE Quality Classified v4 (medium-low): 0.65
  \item Aleph Alpha 6 CC DE Quality Classified (high): 0.17
  \item Aleph Alpha 6 CC DE Quality Classified (medium-high): 0.19
  \item Aleph Alpha 6 CC DE Quality Classified (medium): 0.16
  \item Aleph Alpha 6 CC DE Quality Classified (medium-low): 0.16
  \item FineWeb2-DE Synthetic Rephrase Quality Classified (high): 1.20
  \item FineWeb2-DE Synthetic Rephrase Quality Classified (medium-high): 1.10
  \item FineWeb2-DE Synthetic Rephrase Quality Classified (medium): 0.93
  \item FineWeb2-DE Synthetic Rephrase Quality Classified (medium-low): 0.64
  \item High-Quality-All DE v3: 0.65
  \item Paracrawl Trans EN-DE (shuffled): 0.05
  \item Algebraic Stack: 1.50
  \item DeepMind Maths v2: 2.00
  \item Orca Math Word Problems: 1.50
  \item StarCoder Splitup (all): 10.0
  \item StarCoder Splitup Python Quality Buckets (high): 6.0
  \item StarCoder Python Synthesized (2025-04-01): 4.0
  \item Code Contests: 2.89
  \item APPS: 0.01
  \item Rosetta Code \citep{nanz2014rosettacode}: 0.10
\end{enumerate}

\section{Binary Sequence Metrics} \label{section:binary_sequence_metrics}
\textbf{Normalized gap entropy ($H_g$).}
Let $\{g_i\}_{i=1}^{M-1}$ denote the distances between consecutive boundary positions (i.e., inter-boundary gaps). We define the normalized gap entropy
\[
H_g \;=\; - \frac{1}{\log K} \sum_{k=1}^{K} p(g=k)\,\log p(g=k),
\]
where $p(g=k)$ is the empirical distribution over observed gap lengths and $K$ is the number of distinct gap values. $H_g \in [0,1]$ measures the diversity and unpredictability of boundary spacing: low values indicate constrained or structured spacing, while high values correspond to highly variable gap patterns.

\textbf{CUSUM range ($R_{\mathrm{CUSUM}}$).}
To assess stationarity of the boundary rate over time, we compute the cumulative sum (CUSUM) of the mean-centered boundary sequence,
$S_t = \sum_{i=1}^{t} (b_i - \bar b)$,
where $\bar b$ is the global mean boundary rate. The CUSUM range is then defined as
\[
R_{\mathrm{CUSUM}} = \max_t S_t - \min_t S_t.
\]
For a stationary boundary process, $R_{\mathrm{CUSUM}}$ grows sublinearly with sequence length, whereas large values indicate systematic drift or regime changes in boundary density.

\textbf{Runs test statistic ($Z_{\mathrm{runs}}$).}
We quantify temporal dependence in the boundary sequence using the Wald--Wolfowitz runs test. Let $R$ be the number of runs (maximal contiguous blocks of identical values) in $b_{1:t}$. Under an i.i.d. Bernoulli assumption with the same mean, the expected number of runs $\mu_R$ and variance $\sigma_R^2$ are known in closed form. We report the standardized statistic
\[
Z_{\mathrm{runs}} = \frac{R - \mu_R}{\sigma_R},
\]
where large positive values indicate excessive clustering (long runs of zeros), while values near zero are consistent with independent boundary decisions.

\begin{figure*}
\includegraphics[width=.95\linewidth]{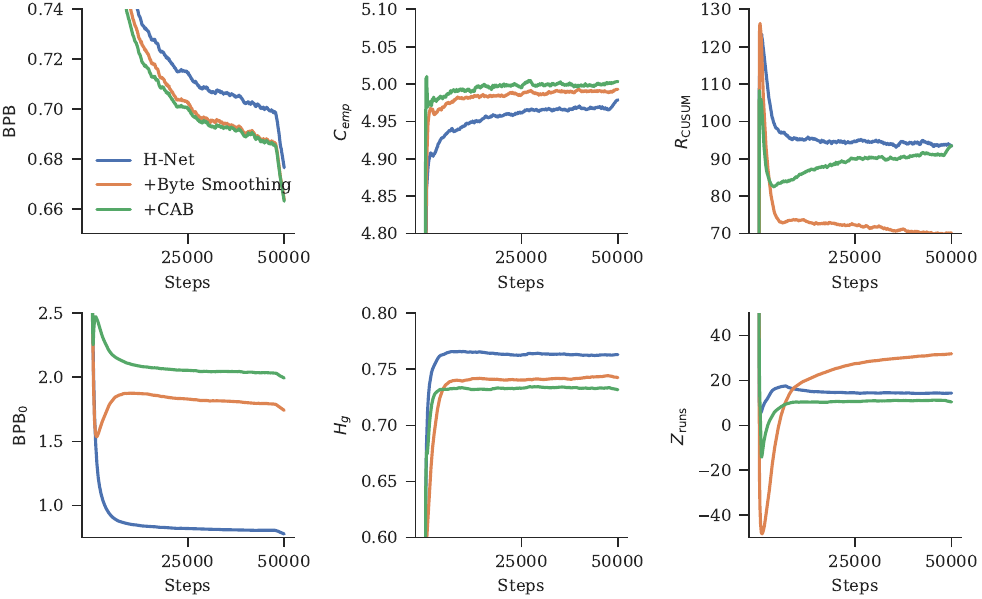}
\caption{Training curves of selected metrics for different configurations of the design ladder. BPB$_0$ indicates the bits-per-byte averaged only over the first bytes of chunks, where higher indicates boundaries on more difficult bytes.}
\label{figure:training_curves}
\end{figure*}

\section{Chunk Illustrations}\label{subsection:chunk_illustration}
We illustrate learned chunk boundaries on a range of within-domain sequences for H-Net (trained by us) and \Sombrero below. Characters that form the start of a new chunk are colored in red. We replace spaces `` '' by underscores ``\_'' in the illustration to make chunk starts on spaces visible. Note that a chunk starting on a character indicates that the prediction of this character was still part of the last chunk, but the prediction of the next character is part of the new chunk (with additional backbone compute). That is: the chunking is shown on the input sequence and not the output sequence. For instance, a chunk start at a whitespace is indicative that a method allocates additional compute for predicting the first character of a new word, while the whitespace itself is still predicted based on the chunk corresponding to the  prior word.

\textbf{Natural Language}\\
\definecolor{splitcolor}{RGB}{255,0,0}
\newcommand{\splitchar}[1]{{\textcolor{splitcolor}{#1}}}
\newcommand{\splitspace}{{\ttfamily\textcolor{splitcolor}{\_}}}

\noindent
\emph{H-Net}: \\ \texttt{Sombrero\_\splitchar{l}earns\_\splitchar{s}\splitchar{e}gmentation\_\splitchar{b}oundaries\_i\splitchar{n}\_\splitchar{h}ierarchical\_\splitchar{s}\splitchar{e}quence\_\splitchar{m}odels\_\splitchar{f}or\_\splitchar{t}okenizer\splitchar{-}\splitchar{f}ree\_\splitchar{b}yte\_s\splitchar{t}reams\splitchar{.}}\\
\emph{Sombrero}: \\ \texttt{S\splitchar{o}mbrero\splitspace{}learns\splitspace{}segmentation\splitspace{}boundaries\splitspace{}in\splitspace{}hierarchical\splitspace{}sequence\splitspace{}model\splitchar{s}\splitspace{}for\splitspace{}tokenizer\splitchar{-}free\splitspace{}byte\splitspace{}streams\splitchar{.}}

\noindent
\emph{H-Net}: \\ \texttt{Sombrero\_\splitchar{l}ern\splitchar{t}\_\splitchar{S}egmentierungs\splitchar{g}renzen\_\splitchar{i}\splitchar{n}\_\splitchar{h}ierarchischen\_\splitchar{S}equenz\splitchar{m}odellen\_\splitchar{f}ür\_\splitchar{t}okenizer\splitchar{-}\splitchar{f}reie\_\splitchar{B}yte\splitchar{f}olgen\splitchar{.}}\\
\emph{Sombrero}: \\ \texttt{S\splitchar{o}mbrero\splitspace{}lernt\splitspace{}\splitchar{S}egmentierung\splitchar{s}grenzen\splitspace{}in\splitspace{}hierarchischen\splitspace{}\splitchar{S}equenz\splitchar{m}odellen\splitspace{}für\splitspace{}tokenizer\splitchar{-}freie\splitspace{}\splitchar{B}yte\splitchar{f}olgen\splitchar{.}}

On natural language (German or English), we qualitatively observe that  \Sombrero prefers aligning chunk boundaries with whitespace, indicating that predicting the first byte of a new word is associated with additional backbone compute. In contrast, H-Net tends to align boundaries with the first character of a word, indicating that the additional compute is used for predicting the second byte, which is counterintuitive and explains the low value of H-Net under the boundary enrichment metric.

\textbf{Simple Sequences}\\

\noindent
\emph{H-Net}{\ttfamily : \\ 1234\splitchar{5}678\splitchar{9}123\splitchar{4}567\splitchar{8}9\splitchar{1}234\splitchar{5}67\splitchar{8}9\splitchar{1}234\splitchar{5}67\splitchar{8}9\splitchar{1}234\splitchar{5}6789\splitchar{1}345\splitchar{6}78\splitchar{9}123\splitchar{4}567\splitchar{8}9}\\
\emph{Sombrero}{\ttfamily : \\ 1\splitchar{2}\splitchar{3}4\splitchar{5}678\splitchar{9}1\splitchar{2}34\splitchar{5}678\splitchar{9}12345678\splitchar{9}12345678\splitchar{9}12345678\splitchar{9}134\splitchar{5}\splitchar{6}\splitchar{7}8\splitchar{9}12\splitchar{3}45678\splitchar{9}}

\noindent
\emph{H-Net}{\ttfamily : \\ abcdef\splitchar{g}hi\splitchar{j}klmno\splitchar{p}\splitchar{q}rstuvw\splitchar{x}yz\splitchar{A}\splitchar{B}CDEF\splitchar{G}HI\splitchar{J}KLMNO\splitchar{P}\splitchar{Q}RSTUVW\splitchar{X}YZ\splitchar{a}\splitchar{b}cd\splitchar{f}ghi\splitchar{j}klmno\splitchar{p}\splitchar{q}rstuvw\splitchar{x}yz}\\
\emph{Sombrero}{\ttfamily : \\ a\splitchar{b}\splitchar{c}\splitchar{d}efghijklmnopqrstuvwxy\splitchar{z}\splitchar{A}\splitchar{B}CDEFGHIJKLMNOPQRSTUVWXY\splitchar{Z}abcd\splitchar{f}ghijklmnopqrstuvwxy\splitchar{z}}

On sequences with simple regularities, we observe that \Sombrero allocates more compute early in the sequence on the first characters but captures regularities better: it identifies that the number sequence is a repetition of ``123456789'' and does not allocate additional compute inside it, except when there is a number missing (like the 2 in the 6-th repetition). \Sombrero also quickly identifies that the character sequence is the alphabet and does not allocate additional compute within the alphabet, except for the third variant where the letter ``e'' is missing.

\textbf{Random}\\

\noindent
\emph{H-Net}{\ttfamily : \\ k\splitchar{Q}7\splitchar{a}\splitchar{Z}P2\splitchar{m}\splitchar{E}9R4x\splitchar{H}8\splitchar{d}\splitchar{L}JYc\splitchar{W}1Tg\splitchar{S}0u\splitchar{N}fM5\splitchar{A}6\splitchar{K}Bev\splitchar{U}oXy3\splitchar{h}\splitchar{C}Fi\splitchar{I}p\splitchar{V}w\splitchar{D}srn\splitchar{q}Otl\splitchar{b}\splitchar{G}8j4\splitchar{a}\splitchar{Z}QY2\splitchar{E}6m\splitchar{P}7x9\splitchar{R}0Wc5\splitchar{H}fU1\splitchar{K}Ad\splitchar{T}JS\splitchar{V}n\splitchar{L}Me\splitchar{B}Xoy}\\
\emph{Sombrero}{\ttfamily : \\ k\splitchar{Q}\splitchar{7}\splitchar{a}\splitchar{Z}P2\splitchar{m}\splitchar{E}9\splitchar{R}4\splitchar{x}H8\splitchar{d}\splitchar{L}JYc\splitchar{W}1\splitchar{T}g\splitchar{S}0u\splitchar{N}f\splitchar{M}5\splitchar{A}6\splitchar{K}Bev\splitchar{U}o\splitchar{X}y3h\splitchar{C}Fi\splitchar{I}p\splitchar{V}w\splitchar{D}srnq\splitchar{O}tl\splitchar{b}\splitchar{G}8j4a\splitchar{Z}QY2\splitchar{E}\splitchar{6}m\splitchar{P}7x\splitchar{9}R0\splitchar{W}c5\splitchar{H}f\splitchar{U}1\splitchar{K}Ad\splitchar{T}JSVn\splitchar{L}Me\splitchar{B}Xoy}

On a random sequence, \Sombrero and H-Net show a very similar pattern, which is somewhat surprising given that they behave very differently on non-random sequences.

\textbf{Retrieval}\\

\noindent
\emph{H-Net}{\ttfamily : \\ 1\splitchar{C}2\splitchar{D}1\splitchar{C}3F2\splitchar{D}1\splitchar{C}4\splitchar{G}3\splitchar{F}2D}\\
\emph{Sombrero}{\ttfamily : \\ 1\splitchar{C}2\splitchar{D}1\splitchar{C}3\splitchar{F}2\splitchar{D}1\splitchar{C}4\splitchar{G}3\splitchar{F}2\splitchar{D}}

On a retrieval task (number to character), we observe that both methods spend additional compute for the prediction of the number, which is infeasible on this sequence as the next number does not follow a pattern, while the character is predictable from the number. This indicates a common failure mode: compute should only be allocated to prediction tasks that are hard, yet feasible.

\textbf{Spaces}\\

\noindent
\emph{H-Net}: \\ \texttt{A\splitchar{B}\_\splitspace{}\splitchar{A}\splitchar{B}\_\_\splitchar{A}\splitchar{B}\_\_\splitchar{A}\splitchar{b}\_\_\splitchar{A}\splitchar{B}}\\
\emph{Sombrero}: \\ \texttt{A\splitchar{B}\splitspace{}\splitspace{}\splitchar{A}\splitchar{B}\splitspace{}\_\splitchar{A}\splitchar{B}\splitspace{}\_\splitchar{A}b\splitspace{}\_\splitchar{A}\splitchar{B}}

\noindent
\emph{H-Net}: \\ \texttt{A\splitchar{B}\_\splitspace{}\_\splitchar{A}\splitchar{B}\_\_\_\splitchar{A}\splitchar{B}\_\_\_\splitchar{A}B\_\_\_\splitchar{A}B}\\
\emph{Sombrero}: \\ \texttt{A\splitchar{B}\splitspace{}\splitspace{}\splitspace{}\splitchar{A}\splitchar{B}\splitspace{}\_\_\splitchar{A}\splitchar{B}\splitspace{}\_\_\splitchar{A}\splitchar{B}\splitspace{}\_\_\splitchar{A}\splitchar{B}}

\noindent
\emph{H-Net}: \\ \texttt{A\splitchar{B}\_\splitspace{}\_\_\splitchar{A}\splitchar{B}\_\_\_\_\splitchar{A}\splitchar{B}\_\_\_\_\splitchar{A}B\_\_\_\_\splitchar{A}B}\\
\emph{Sombrero}: \\ \texttt{A\splitchar{B}\splitspace{}\splitspace{}\splitspace{}\_\splitchar{A}\splitchar{B}\splitspace{}\_\_\_\splitchar{A}\splitchar{B}\splitspace{}\_\_\_\splitchar{A}\splitchar{B}\splitspace{}\_\_\_\splitchar{A}\splitchar{B}}

We note that \Sombrero tends to split on all spaces, unless the context suggests that a certain number of spaces occur in a group, in which case \Sombrero splits only on the first space of a group and the first non-space characters.

\textbf{Observations}\\

Overall, we observe that \Sombrero always places a chunk boundary on the second character, indicating that chunking is not purely content- but also position-dependent.


\end{document}